\documentclass{article} 
\usepackage{iclr2025_conference,times}

\usepackage{amsmath,amsfonts,bm}

\def\eqref#1{equation~\ref{#1}}

\def\1{\bm{1}}

\DeclareMathAlphabet{\mathsfit}{\encodingdefault}{\sfdefault}{m}{sl}
\SetMathAlphabet{\mathsfit}{bold}{\encodingdefault}{\sfdefault}{bx}{n}

\usepackage{hyperref}
\usepackage{graphicx}
\usepackage{booktabs}
\usepackage{multirow}
\usepackage[table]{xcolor}
\usepackage{amsmath, amsfonts}
\usepackage{subcaption}
\usepackage{makecell}
\usepackage{algorithm}
\usepackage{algorithmic}
\usepackage{url}
\usepackage{wrapfig}

\title{TetherCache: Stabilizing Autoregressive Long-Form Video Generation with Gated Recall and Trusted Alignment}

\author{Yu Meng\textsuperscript{\rm 1},
Xiangyang Luo\textsuperscript{\rm 1},
Letian Li\textsuperscript{\rm 1},
Wenyuan Jiang\textsuperscript{\rm 2}, 
Chen Gao\textsuperscript{\rm 1},
Xinlei Chen\textsuperscript{\rm 1}, \\
\textbf{Yong Li}\textsuperscript{\rm 1},
\textbf{Xiao-Ping Zhang}\textsuperscript{\rm 1}  \\
\textsuperscript{\rm 1}Tsinghua University, \textsuperscript{\rm 2}D-INFK, ETH Zürich \\
\texttt{mengy24@mails.tsinghua.edu.cn} \\
}

\iclrfinalcopy 
\begin{document}

\maketitle

\begin{figure}[ht]
    \centering
        \includegraphics[width=0.95\linewidth]{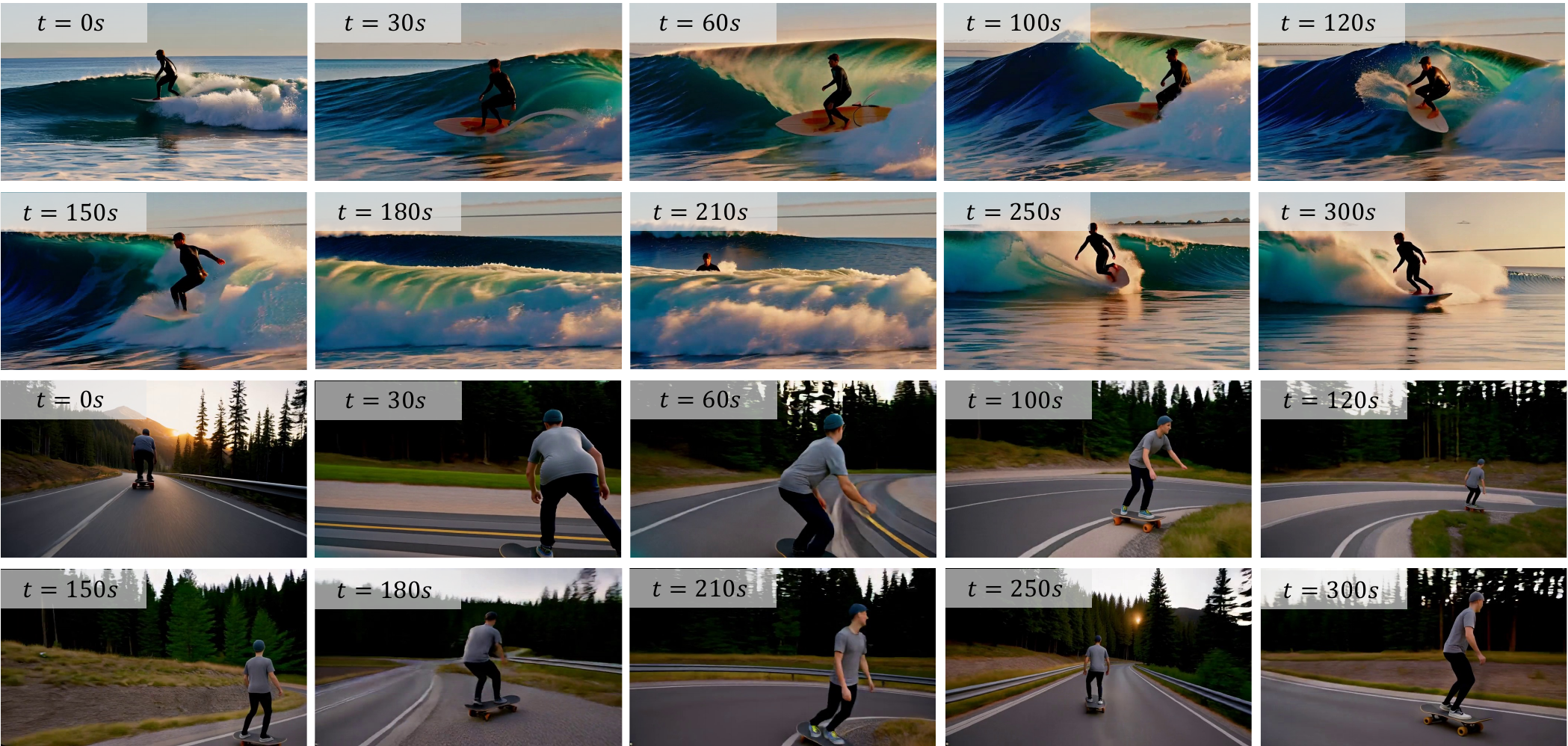}
    \caption{\textbf{TetherCache enables ultra-long video generation for autoregressive diffusion models.} Without requiring any additional model training, TetherCache maintains visual quality even for 5-minute video generation.}
    \label{fig:teaser}
\end{figure}

\begin{abstract}
Autoregressive video diffusion models provide a natural formulation for streaming and variable-length video generation by conditioning newly generated frames on previously generated content. However, extending these models to minute-level generation remains challenging: the limited KV-cache budget prevents the model from retaining the full history, while repeatedly conditioning on self-generated frames induces a context distribution shift that accumulates over time, leading to visual artifacts, quality degradation, and temporal drift. In this paper, we propose \textbf{TetherCache}, a training-free and plug-and-play cache management strategy for drift-resistant long video generation. TetherCache organizes the cache into sink, memory, and recent regions, and introduces two complementary mechanisms. First, \textbf{GRAB} (\textbf{G}ated \textbf{R}ecall with \textbf{A}ttention-Diversity \textbf{B}alancing) selects long-range memory frames using a gated score that combines attention-based relevance with temporal diversity, preserving informative yet diverse historical context under a fixed cache budget. Second, \textbf{TAME} (\textbf{T}rusted \textbf{A}lignment via \textbf{M}emory \textbf{E}diting) lightly edits newly recalled memory tokens by aligning their statistics to a trusted context distribution, reducing the pollution caused by drifted historical features. Built on Self-Forcing, TetherCache consistently improves long-video generation quality on VBench-Long across 30s, 60s, and 240s settings. In particular, for 240s generation, it substantially improves overall and semantic scores while reducing quality drift from 7.84 to 1.33, demonstrating its effectiveness for stable long-horizon autoregressive video diffusion. Code and demos can be found at \url{https://my4f175.github.io/TetherCache}.

\end{abstract}

\section{Introduction}

Recent advances in diffusion-based video generation have substantially improved the visual fidelity and temporal coherence of synthesized videos.
Despite this progress, most existing models remain optimized for short clips of only dozens of frames, falling short of the requirements of emerging applications such as game simulation~\citep{wang2026matrixgame30realtimestreaming,tang2026hunyuangamecraft2instructionfollowinginteractivegame}, world modeling~\citep{seo2026groundingworldsimulationmodels,sun2025worldplaylongtermgeometricconsistency,nvidia2026worldsimulationvideofoundation}, and interactive cinema~\citep{yang2025longlive,luo2026filmweaver,shin2026motionstreamrealtimevideogeneration}, where models are expected to continuously synthesize minute-level video streams.
Generating such long-form videos is challenging because the model must preserve both local motion continuity and long-range semantic consistency while avoiding the progressive degradation.

Autoregressive (AR) video diffusion models provide a natural formulation for this setting by generating video frames sequentially.
Given previously generated frames or chunks (for simplicity, we use “frame” hereafter to uniformly refer to either a frame or a chunk) as conditions, an AR model predicts the next frame and can therefore be rolled out to arbitrary lengths without changing the underlying generation interface.
During inference, historical frames are typically stored in the KV cache and reused as conditioning context for future generation, enabling efficient and temporally coherent video synthesis.
However, when AR diffusion models are extended far beyond their training horizon, this seemingly straightforward rollout exposes two fundamental challenges:
\begin{enumerate}
    \item \textbf{Context Length Limitation}: The amount of historical context grows linearly with the generated video length, whereas the KV-cache budget is fixed by memory and latency constraints. The model must therefore discard part of the history, which can remove useful long-range cues and weaken semantic consistency.
    \item \textbf{Context Distribution Shift}: During training, the model is conditioned on ground-truth frames or short self-generated rollouts whose distribution remains close to the training context. During long-video inference, however, the model must repeatedly condition on its own generated outputs far beyond the training window. The resulting mismatch accumulates over time and manifests as visual artifacts, color shifts, noise amplification, and positional or semantic drift.
\end{enumerate}

These two challenges are coupled but require different remedies.
A cache policy should retain historical information that is useful for future generation under a constrained budget, but it should also prevent drifted historical features from degrading the conditioning context.
Naively keeping the most recent frames preserves short-term continuity but loses long-range information, while blindly recalling generated frames may reintroduce statistically unreliable features generated after substantial rollout drift.
This suggests that long-form AR video generation requires cache management that is both selective and distribution-aware.
Existing context management methods~\citep{yesiltepe2026infinityropeactioncontrollableinfinitevideo,yi2025deep} often select or compress context according to a single criterion, such as recency or attention, and therefore do not explicitly balance relevance with diversity.
Moreover, although sink frames are widely used as persistent context~\citep{ye2026dysinkdynamicframesinks,yi2025deep,li2026rolling}, their role is typically limited to stabilizing attention or positional extrapolation. The reliable distributional priors carried by these early high-quality frames remain under-exploited for repairing drifted historical features.

To this end, we propose \textbf{TetherCache}, a training-free and plug-and-play cache management strategy for drift-resistant long video generation.
TetherCache organizes the KV cache into three regions: \textbf{Sink}, \textbf{Recent}, and \textbf{Memory}.
The Sink cache stores several initial frames as trusted sink frames, motivated by the observation that early AR outputs are generated under contexts closest to those seen during training and usually have higher quality.
The Recent cache keeps the latest frames to preserve local temporal coherence.
The Memory cache stores a compact subset of intermediate historical frames that supplies long-range context under a fixed cache budget.
By separating these roles, TetherCache provides a simple structure in which stable sink frames, local continuity, and long-range recall can be handled explicitly.

Within this structure, we introduce two complementary mechanisms.
First, \textbf{GRAB} (Gated Recall via Attention-Diversity Balancing) addresses the context length limitation by deciding which historical frames deserve memory slots.
GRAB scores each candidate using both attention-based relevance and temporal diversity: the former estimates how useful a frame is for the current rollout, while the latter discourages the memory from being filled with redundant neighboring frames.
This gated selection preserves informative yet diverse long-range context instead of relying on a purely recency-based cache.
Second, \textbf{TAME} (Trusted Alignment via Memory Editing) addresses context distribution shift by lightly editing newly admitted memory tokens.
Using the trusted cache region as a reference, TAME aligns the statistics of admitted KV tokens toward a more reliable context distribution, reducing the risk that drifted historical features pollute future attention computation.
Together, GRAB and TAME allow TetherCache to recall useful history while tethering the recalled context to a stable distribution.

We implement TetherCache on top of Self-Forcing~\citep{NEURIPS2025_f4823f83} without retraining or modifying model parameters.
Experiments on VBench-Long~\citep{huang2023vbench} show consistent improvements across 30s, 60s, and 240s generation settings.
In the challenging 240s setting, TetherCache substantially improves overall and semantic scores, and reduces $\Delta$ Quality Drift from 7.84 to 1.33 compared with the Self-Forcing baseline.
Ablation studies further show that GRAB and TAME contribute complementary gains: GRAB improves long-range recall and reduces drift, while TAME further stabilizes the cached context and improves final generation quality.

Our contributions are summarized as follows:
\begin{enumerate}
    \item We identify two coupled cache challenges in long-form AR video diffusion: limited historical coverage under a fixed KV-cache budget and context distribution shift caused by long-horizon self-conditioning.
    \item We propose TetherCache, a training-free cache management framework that jointly addresses the two failures through GRAB, which performs relevance-diversity memory recall, and TAME, which exploits sink-frame distributional priors for trusted statistic alignment.
    \item We demonstrate that TetherCache improves long-video generation quality on VBench-Long across multiple durations, with especially strong gains in 240s generation and quality-drift reduction, without modifying model parameters or the original KV-cache representation.
\end{enumerate}
\section{Related Work}

\subsection{AR Video Diffusion}

The AR generation paradigm naturally aligns with the causal structure of video streams. To achieve AR video generation, one line of work~\citep{gao2025ca2,hu2026acdit,ICLR2025_3ab228c4} applies Teacher Forcing~\citep{6795228}, training diffusion models to predict subsequent video frames conditioned on clean historical frames. Another line of work~\citep{chen2025skyreelsv2infinitelengthfilmgenerative,gu2025long,ai2025magi1autoregressivevideogeneration,pmlr-v267-song25b} builds on Diffusion Forcing~\citep{chen2025diffusion}, enabling models to learn to denoise frames with independently sampled noise levels. CausVid~\citep{yin2025causvid} first proposes distilling a bidirectional diffusion model into an AR generative model using a holistic distribution-level loss~\citep{yin2024onestep}. Based on this idea, Self-Forcing~\citep{NEURIPS2025_f4823f83} mitigates the train–test discrepancy in AR generation by introducing KV-Cache and Self-Rollout during training. Causal Forcing~\citep{zhu2026causal} further improves the initialization process of Self-Forcing, achieving higher generation quality. However, although these methods enable AR generation, when extrapolating to long videos beyond the training window length, they still suffer from quality drift caused by error accumulation, facing the challenge of exposure bias~\citep{DBLP:journals/corr/RanzatoCAZ15}.

\subsection{Mitigating Exposure Bias in Long Video Extrapolation}

To mitigate exposure bias in long video generation, SVI~\citep{li2025stablevideoinfinityinfinitelength} and Helios~\citep{yuan2026heliosrealrealtimelong} explicitly inject generation errors into training context, enabling robust generation even in the presence of quality drift. Self-Forcing++~\citep{cui2025self} extends the training window length, while Rolling Forcing~\citep{liu2025rolling} and HiAR~\citep{zou2026hiarefficientautoregressivelong} slow error accumulation through joint multi-frame denoising. However, all of these methods incur the cost of retraining the model.

Another line of work achieves long-horizon extrapolation or inference acceleration by designing training-free context management strategies, including the use of sink tokens~\citep{yesiltepe2026infinityropeactioncontrollableinfinitevideo,yi2025deep,li2026rolling,kim2026memrope,zhao2026relaxforcingrelaxedkvmemory,ye2026dysinkdynamicframesinks,li2026trainshortinferencelong}, positional encoding designs~\citep{yesiltepe2026infinityropeactioncontrollableinfinitevideo,kim2026memrope,li2026rolling,mao2026packforcingshortvideotraining,li2026trainshortinferencelong}, cache compression and retrieval~\citep{kim2026memrope,wu2026echoforcingscenememoryframework,mao2026packforcingshortvideotraining,ji2026forcingkvhybridkvcache,samuel2026fastautoregressivevideodiffusion}, and efficient attention mechanism~\citep{li2026longhorizonstreamingvideogeneration,samuel2026fastautoregressivevideodiffusion}. Different from prior methods that mainly focus on cache layout or positional encoding, but largely overlook that sink frames themselves carry reliable distributional priors for repairing drifted contexts, TetherCache explicitly leverages the trusted statistics of the sink region to align newly recalled memory tokens, thereby jointly addressing memory selection and context distribution shift under the original KV-cache budget.
\section{Preliminary}

\paragraph{AR video diffusion.}
We consider long-form video generation in an AR setting. A video latent sequence $\mathbf{z}_{1:n}$ is generated from left to right,
\begin{equation}
    p(\mathbf{z}_{1:n}) = \prod_{i=1}^{n} p(\mathbf{z}_i \mid \mathbf{z}_{<i}).
\end{equation}
Each conditional distribution is modeled by a diffusion generator $G_\theta$. At denoising step $j$, the generator receives a noisy latent $\mathbf{z}_i^{t_j}$ and attends to the KV cache produced by previously generated frames,
\begin{equation}
\mathbf{z}_i^{t_{j-1}} =
\Psi\left(
G_\theta\left(\mathbf{z}_i^{t_j}, t_j, \mathbf{K}_{<i}, \mathbf{V}_{<i}\right),
t_{j-1}
\right),
\end{equation}
where $\Psi$ denotes the forward noising operator from level $t_j$ to $t_{j-1}$, and $\mathbf{z}_i^{t_T}\sim\mathcal{N}(0,I)$.

In practice, the cache cannot grow with video length. Let the cache budget be $K$ latent frames and let each latent frame contain $L$ tokens. A standard local-cache baseline stores only the most recent $K$ frames, which bounds memory but removes all earlier context. This is problematic for long rollouts: early sink frames that calibrate attention statistics are discarded, while semantically useful distant frames become inaccessible.
\section{Method}

\begin{figure*}[t!]
    \centering
        \includegraphics[width=\linewidth]{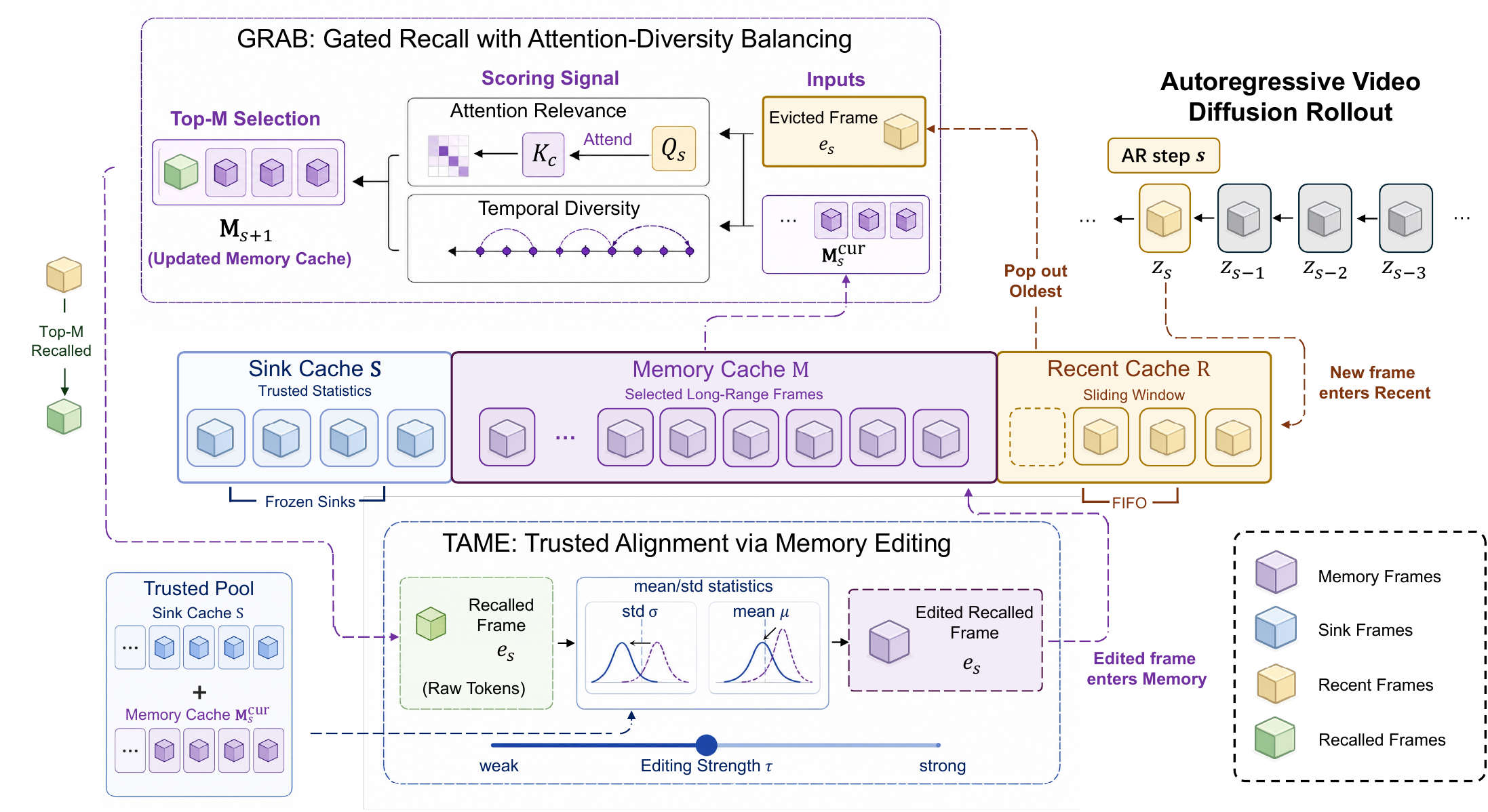}
    \caption{\textbf{Overview of TetherCache.}
TetherCache divides the fixed KV cache into Sink, Memory, and Recent regions. GRAB performs relevance-diversity memory recall from evicted recent frames and existing memory, while TAME uses trusted sink statistics to align newly recalled KV tokens.}
\label{fig:tethercache-framework}
\end{figure*}

\subsection{Overview}

We propose \textbf{TetherCache}, a training-free, drop-in cache management strategy for stabilizing long-form AR video generation. As shown in Figure~\ref{fig:tethercache-framework}, TetherCache keeps the original cache tensor shape unchanged, but reinterprets its $K$ frame slots as three contiguous regions:
\begin{equation}
    \underbrace{\mathbf{S}}_{S\ \text{sink frames}}
    \;\Vert\;
    \underbrace{\mathbf{M}}_{M\ \text{memory frames}}
    \;\Vert\;
    \underbrace{\mathbf{R}}_{R\ \text{recent frames}},
    \label{eq:tether-layout}
\end{equation}
and $K = S + M + R$. Here $\mathbf{S}$ is a frozen sink region, $\mathbf{M}$ is a selective long-range memory, and $\mathbf{R}$ is a short sliding window. During warm-up, since the cache has not yet been filled, newly generated frames are added to the cache unconditionally. Once the cache first fills, the first $S$ frames are frozen as sink frames, the next $M$ frames initialize memory, and the last $R$ frames form the recent window.

After warmup, every newly generated frame is written to the tail of $\mathbf{R}$. The front of $\mathbf{R}$ is evicted and becomes a candidate for long-range storage. TetherCache then performs two operations: \textbf{GRAB} decides whether the evicted frame should be recalled into $\mathbf{M}$, and \textbf{TAME} edits the recalled K/V tokens before they are written into memory.

TetherCache is applied independently in each attention layer. It does not allocate a parallel KV buffer: the memory region is the same physical slice of the model's original $\mathbf{K}$ and $\mathbf{V}$ cache. We only maintain lightweight metadata, including the global frame index of each memory slot and frozen sink statistics. To avoid corrupting rotary positions when a frame is moved into memory, TetherCache stores unrotated keys and applies block-relative RoPE when reading from the cache.

\subsection{GRAB: Gated Recall with Attention-Diversity Balancing}

GRAB determines which frames deserve the limited memory slots. Suppose that at AR rollout step $s$, the current memory contains $m\le M$ frames and the recent window evicts one frame $e_s$. GRAB forms the candidate pool
\begin{equation}
    \mathcal{P}_s = \mathbf{M}_{s}^{\mathrm{cur}} \cup \{e_s\}.
    \label{eq:grab-pool}
\end{equation}
It then re-scores every candidate in $\mathcal{P}_s$, including frames already stored in memory. The new memory is the top-$M$ subset under a score combining attention-based importance and temporal diversity:
\begin{equation}
    \phi(c) = \phi^{\mathrm{imp}}(c) + \alpha\,\phi^{\mathrm{div}}(c),
    \qquad c\in\mathcal{P}_s,
    \label{eq:grab-score}
\end{equation}
where $\alpha$ controls the strength of the diversity term. The evicted frame is recalled iff it enters this top-$M$ set, otherwise it is dropped. Since old memory entries are also re-evaluated, GRAB is not append-only: stale memories can be demoted when a more relevant evicted frame appears.

\paragraph{Attention-mass importance.}
Let $\mathbf{Q}_s\in\mathbb{R}^{L\times H\times D}$ be the clean queries of the current frame, where $L$ is the number of tokens per latent frame, $H$ is the number of heads, and $D$ is the head dimension. For a candidate frame $c$ with key tokens $\mathbf{K}_c\in\mathbb{R}^{L\times H\times D}$, GRAB estimates how much attention mass the candidate would attract from the current frame:
\begin{equation}
    \ell(c) =
    \frac{1}{HL^2}\sum_{h=1}^{H}\sum_{q=1}^{L}\sum_{k=1}^{L}
    \frac{\left\langle \mathbf{Q}_{s,q}^{h}, \mathbf{K}_{c,k}^{h}\right\rangle}{\sqrt{D}}, \quad
    \phi^{\mathrm{imp}}(c) =
    \frac{\exp\left(\ell(c)\right)}
    {\sum_{c'\in\mathcal{P}_s}\exp\left(\ell(c')\right)}.
    \label{eq:importance}
\end{equation}
This score is computed from the same query-key geometry used by attention, but pooled to one scalar per candidate frame. The implementation also supports a lightweight equivalent that sums queries before the matrix product, avoiding explicitly constructing the full $L\times |\mathcal{P}_s|L$ attention tensor.

\paragraph{Temporal diversity.}
Importance alone can fill memory with adjacent frames from the same short interval. GRAB therefore adds a temporal diversity bonus that favors candidates covering different parts of the rollout. Let $g_c\in\mathbb{Z}$ be the global frame index of candidate $c$, and set
\begin{equation}
    \sigma_s = \max\left(1, \frac{1}{2}\left(\max_{c\in\mathcal{P}_s}g_c - \min_{c\in\mathcal{P}_s}g_c + 1\right)\right).
\end{equation}
We measure the redundancy of $c$ as its maximum importance-weighted temporal similarity to any other candidate:
\begin{equation}
    r(c) = \max_{c'\neq c}
    \exp\left(-\frac{|g_c-g_{c'}|}{\sigma_s}\right)
    \phi^{\mathrm{imp}}(c').
\end{equation}
The diversity bonus is
\begin{equation}
    \phi^{\mathrm{div}}(c)=\max\left(0, 1-r(c)\right).
    \label{eq:diversity}
\end{equation}
A candidate receives a small bonus if it lies near another already important candidate, and a large bonus if it represents a temporally isolated portion of the video. Finally, GRAB selects
\begin{equation}
    \mathbf{M}_{s+1} = \operatorname{TopM}_{c\in\mathcal{P}_s}\, \phi(c).
    \label{eq:grab-select}
\end{equation}

\subsection{TAME: Trusted Alignment via Memory Editing}

GRAB recalls useful distant frames, but recalled K/V tokens may have drifted statistically during long rollouts. Such drift changes attention calibration: even semantically relevant tokens can distort the softmax distribution if their K/V statistics no longer match the trusted context. TAME addresses this by editing only newly recalled frames, leaving sink frames, recent frames, and already accepted memory entries untouched.

For a recalled frame, let $\mathbf{x}\in\mathbb{R}^{L\times H\times D}$ denote either its key or value tokens. TAME constructs a trusted pool
\begin{equation}
    \mathcal{T}_s = \mathbf{S} \cup \mathbf{M}_{s}^{\mathrm{cur}},
    \label{eq:trusted-pool}
\end{equation}
where sink frames provide a stable distributional tether and existing memory provides slowly evolving rollout context. When $\mathbf{x}$ denotes key or value tokens, $\mathcal{T}_s$ denotes the corresponding key or value tokens from the trusted pool. K and V are edited separately. We compute per-head, per-channel statistics over token positions,
\begin{equation}
    (\mu_{\mathbf{x}}, \sigma_{\mathbf{x}})=\operatorname{Stat}(\mathbf{x}),
    \qquad
    (\mu_{\mathcal{T}}, \sigma_{\mathcal{T}})=\operatorname{Stat}(\mathcal{T}_s),
\end{equation}
where $\mu,\sigma\in\mathbb{R}^{H\times D},$ and $\operatorname{Stat}$ returns the mean and standard deviation over the batch/token axes. TAME then performs a partial alignment:
\begin{equation}
    \tilde{\mathbf{x}} =
    \sigma_{\mathcal{T}}\odot
    \frac{\mathbf{x}-\mu_{\mathbf{x}}}{\sigma_{\mathbf{x}}}
    + \mu_{\mathcal{T}}, \quad
    \hat{\mathbf{x}} = (1-\tau)\mathbf{x} + \tau\tilde{\mathbf{x}},
    \qquad \tau\in[0,1].
    \label{eq:tame}
\end{equation}
This preserves the normalized content pattern of the recalled frame while aligning its first- and second-order statistics to trusted cache tokens. The edit is conservative: if the candidate is already close to the trusted distribution, then $\tilde{\mathbf{x}}\approx\mathbf{x}$ and the update is nearly identity. If it has drifted, the blend pulls it back proportionally to $\tau$.

\subsection{Putting It Together}

At each steady-state frame, TetherCache executes the following procedure in every attention layer: (1) roll the recent window and snapshot the evicted frame; (2) write the current frame to the recent tail on every denoising pass; (3) run GRAB over existing memory plus the evicted frame; (4) apply TAME if that frame is newly recalled; and (5) rewrite the memory region with the selected top-$M$ frames.

The method preserves the cache budget and attention interface of the baseline. Its overhead is limited to one small candidate scoring operation per generated frame and per-layer metadata for memory indices and sink statistics. By combining short-term recency, long-term selective recall, and trusted statistical alignment, TetherCache stabilizes long-form generation without model finetuning or changing the diffusion sampler.

\section{Experiments}

\subsection{Experimental Setup}

\paragraph{Implementation Details.}

We implement the proposed method upon Self Forcing, an AR diffusion model trained on Wan2.1~\citep{wan2025wanopenadvancedlargescale}. We use a total cache budget of $K=21$ (where $S=3$ and $R=4$) as the default cache configuration to align with the training window length of Self-Forcing, with $\alpha=0.35$ and $\tau=0.6$ as the default hyperparameters. Following ~\citep{yesiltepe2026infinityropeactioncontrollableinfinitevideo,kim2026memrope}, we use block-wise relative RoPE to further maintain the distributional stability of the KV cache.

\paragraph{Evaluation Protocol.} We randomly sample 128 prompts from MovieGen-Bench~\citep{polyak2024movie} to generate videos of 30s and 60s, and sample 32 prompts to generate 240s videos at a resolution of $832 \times 480$, and evaluate them using VBench metrics. All prompts are refined by Qwen2.5-7B-Instruct~\citep{qwen2025qwen25technicalreport}. To evaluate quality drift in long video generation, following~\citep{zhang2025framecontextpackingdrift}, we compute the difference in Imaging Quality between the first and last clips of each video, denoted as $\Delta$ Quality Drift. The $\Delta$ Quality Drift value intuitively reflects the degree of error accumulation. We also conduct a user study follow the Two-Alternative Forced Choice
(2AFC) protocol.

\subsection{Main Results}

\paragraph{Quantitative Results.}

\begin{table}[t]
\centering
\caption{\textbf{Quantitative comparison for 30s, 60s and 240s video generation on VBench-Long.} The \textbf{best} performance is highlighted in bold, and the \underline{second-best} is underlined. \textbf{Cons.} stands for Consistency.}
\resizebox{\textwidth}{!}{
\label{tab:main-results}
\begin{tabular}{lcccccccc}
\toprule
\multirow{2}{*}[-1.4ex]{\textbf{Model}} &
\multirow{2}{*}[-1.4ex]{\makecell[c]{\textbf{Dynamic}\\\textbf{Degree}} $\uparrow$} &
\multirow{2}{*}[-1.4ex]{\makecell[c]{\textbf{Imaging}\\\textbf{Quality}} $\uparrow$} &
\multirow{2}{*}[-1.4ex]{\makecell[c]{\textbf{Color}\\\textbf{Cons.}} $\uparrow$} &
\multirow{2}{*}[-1.4ex]{\makecell[c]{\textbf{Overall}\\\textbf{Cons.}} $\uparrow$} &
\multirow{2}{*}[-1.4ex]{$\Delta$\makecell[c]{\textbf{Quality}\\\textbf{Drift}}$\downarrow$} &
\multicolumn{3}{c}{\textbf{Evaluation Scores} $\uparrow$} \\
\cmidrule(lr){7-9}
& & & & & &
\makecell[c]{\textbf{Total}\\\textbf{Score}} &
\makecell[c]{\textbf{Quality}\\\textbf{Score}} &
\makecell[c]{\textbf{Semantic}\\\textbf{Score}} \\
\midrule
\multicolumn{9}{c}{30 seconds} \\
\midrule
CausVid
& \textbf{52.00} & 67.38 & 78.04 & 24.94 & 4.45 & 81.79 & \textbf{83.42} & 75.31 \\
Self Forcing
& 24.29 & \underline{69.79} & 68.27 & \textbf{26.11} & 0.95 & 80.86 & 82.03 & 76.16 \\
$\infty$-RoPE
& 45.71 & 69.22 & 70.21 & 25.60 & 1.01 & \underline{82.31} & 82.95 & \underline{79.73} \\
MemRoPE
& 41.43 & 69.40 & 72.94 & \underline{26.03} & \underline{0.09} & 81.74 & 82.92 & 77.01 \\
Deep Forcing
& 37.14 & 68.12 & \underline{89.06} & 25.40 & 4.17 & 82.04 & 82.76 & 79.17 \\
\rowcolor{blue!10}
\textbf{Ours}
& \underline{48.57} & \textbf{70.00} & \textbf{89.74} & 25.71 & \textbf{-0.51} & \textbf{82.70} & \underline{83.40} & \textbf{79.94} \\
\midrule
\multicolumn{9}{c}{60 seconds} \\
\midrule
CausVid
& \underline{39.33} & 67.66 & 78.35 & 24.45 & 2.57 & 80.06 & \underline{82.16} & 71.67 \\
Self Forcing
& 34.48 & 67.06 & 67.56 & 23.49 & 11.34 & 78.82 & 81.53 & 67.97 \\
$\infty$-RoPE
& 22.76 & 65.00 & 56.23 & 23.96 & 11.51 & 78.40 & 80.60 & 69.59 \\
MemRoPE
& 34.48 & 67.36 & 81.95 & 24.74 & 4.75 & 79.69 & 81.04 & 74.27 \\
Deep Forcing
& 23.45 & \textbf{70.01} & \underline{90.96} & \textbf{26.69} & \underline{1.97} & \underline{81.49} & 82.01 & \underline{79.42} \\
\rowcolor{blue!10}
\textbf{Ours}
& \textbf{43.45} & \underline{69.84} & \textbf{95.01} & \underline{26.07} & \textbf{0.33} & \textbf{82.11} & \textbf{82.51} & \textbf{80.49} \\
\midrule
\multicolumn{9}{c}{240 seconds} \\
\midrule
CausVid
& 33.75 & \underline{67.31} & 72.21 & 23.42 & 8.03 & 78.45 & \textbf{82.03} & 64.12 \\
Self Forcing
& 28.99 & 58.51 & 78.78 & 14.98 & 7.84 & 71.22 & 77.95 & 44.34 \\
$\infty$-RoPE
& \textbf{39.71} & 56.29 & 78.84 & 18.52 & 8.84 & 72.34 & 79.02 & 45.61 \\
MemRoPE
& 34.03 & 64.69 & 79.21 & 21.77 & 7.94 & 75.63 & 80.13 & 57.62 \\
Deep Forcing
& 25.42 & 65.59 & \underline{90.47} & \underline{24.44} & \underline{3.96} & \underline{78.44} & 79.92 & \underline{72.54} \\
\rowcolor{blue!10}
\textbf{Ours}
& \underline{35.50} & \textbf{68.58} & \textbf{98.96} & \textbf{25.70} & \textbf{1.33} & \textbf{80.17} & \underline{81.50} & \textbf{74.85} \\
\bottomrule
\end{tabular}
}
\end{table}

Table~\ref{tab:main-results} reports quantitative comparisons with several state-of-the-art baselines under 30s, 60s, and 240s generation settings. Overall, TetherCache achieves the best Total Score at all three durations, showing that the proposed cache management strategy consistently improves long-video generation quality without retraining the backbone model. In the 30s setting, our method obtains the highest Total Score and Semantic Score, while also achieving the lowest $\Delta$ Quality Drift. This indicates that TetherCache does not sacrifice short-horizon generation quality despite introducing long-range memory management.

The advantage becomes more pronounced as the generation length increases. For 60s videos, TetherCache achieves the best Dynamic Degree, Quality Score, Semantic Score, Total Score, and $\Delta$ Quality Drift, and remains competitive on Imaging Quality and Overall Consistency. In the more challenging 240s setting, most baselines suffer from substantial degradation: Self-Forcing drops to 58.51 Imaging Quality and 44.34 Semantic Score, while its $\Delta$ Quality Drift reaches 7.84. In contrast, TetherCache maintains the best Imaging Quality, Overall Consistency, Total Score, and Semantic Score, and reduces $\Delta$ Quality Drift to 1.33. These results demonstrate that retaining relevant and diverse historical context with GRAB and stabilizing admitted memory tokens with TAME effectively mitigates both semantic drift and visual quality degradation during long AR rollouts.

\paragraph{Qualitative Results.}

\begin{figure}[t]
    \centering
        \includegraphics[width=\linewidth]{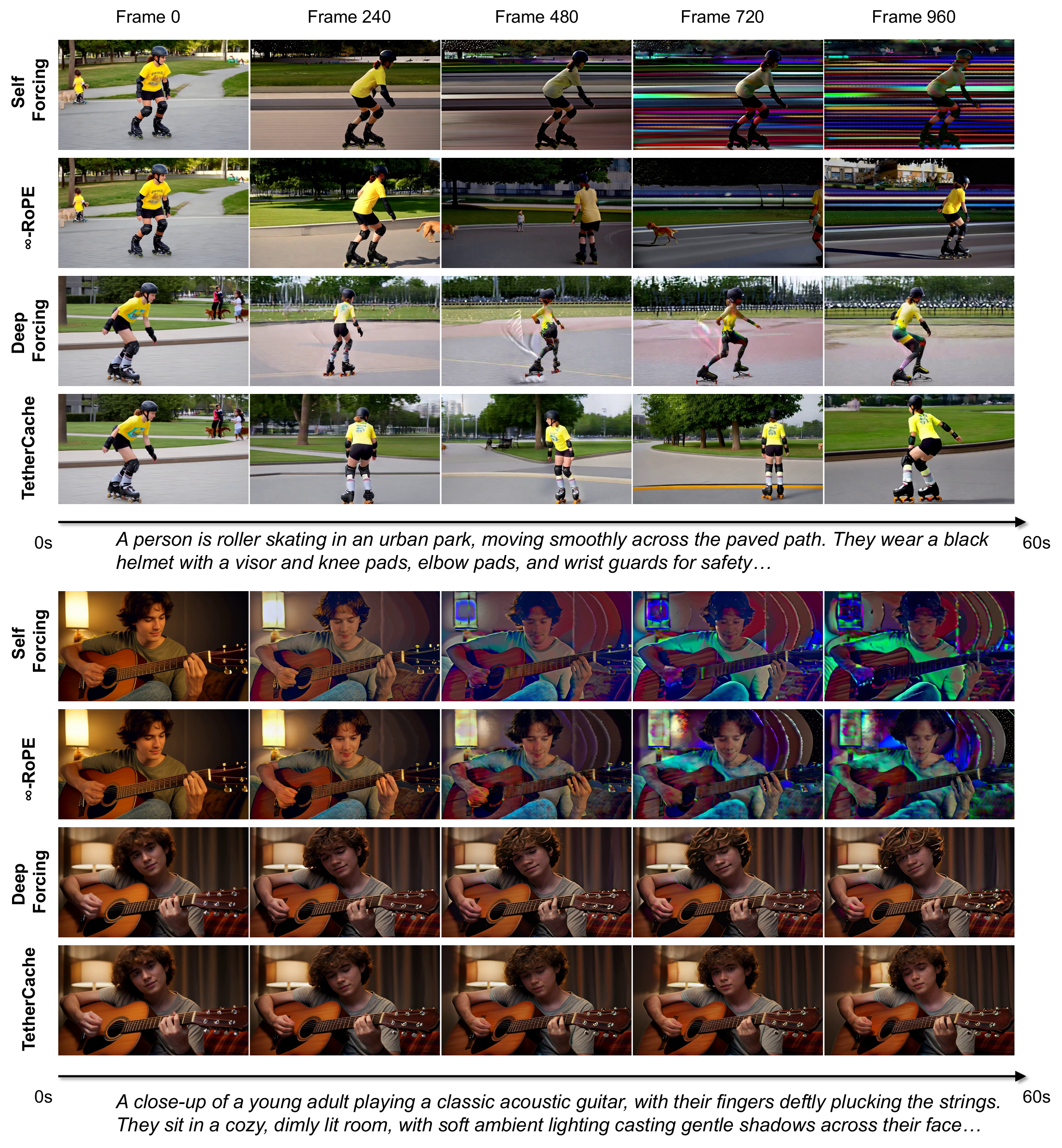}
    \caption{\textbf{Qualitative comparisons with baselines.} TetherCache better preserves visual quality throughout long rollouts, while suppressing accumulated artifacts in later clips.}
    \label{fig:qualitative-results}
\end{figure}

Figure~\ref{fig:qualitative-results} presents qualitative comparisons between our method and representative baselines in long-video generation. Baseline methods tend to suffer from visible degradation as the rollout becomes longer, including accumulated noise, color or illumination shifts, distorted object shapes, and weakened semantic consistency. These artifacts are especially pronounced in later clips, where the model has repeatedly conditioned on its own imperfect generations. In contrast, our method maintains more stable visual quality and preserves the main scene layout, object identity, and motion continuity. The improvement suggests that TetherCache not only enhances quantitative scores, but also effectively suppresses perceptible drift in actual generated videos. By recalling informative historical frames and regularizing recalled memory tokens, we provide a cleaner and more reliable conditioning context throughout the AR rollout.

\paragraph{User Studies.}

\begin{table}[h]
\centering
\caption{\textbf{User Preference Rates.} Pairwise 2AFC user preference percentages of TetherCache over each baseline.}
\label{tab:user-studies}
\resizebox{0.6\textwidth}{!}{
\begin{tabular}{l|cccc}
\toprule
\textbf{Method} & 
\makecell{\textbf{Subject}\\\textbf{Consistency}} &
\makecell{\textbf{Imaging}\\\textbf{Quality}} &
\makecell{\textbf{Overall}\\\textbf{Quality}} &
\textbf{Avg.}\\
\midrule
vs. CausVid
& 89.29 & 100.00 & 96.43 & 95.24 \\
vs. Self Forcing
& 71.43 & 71.43 & 67.86 & 70.24 \\
vs. $\infty$-RoPE
& 71.43 & 67.86 & 67.86 & 69.05 \\
vs. MemRoPE
& 71.43 & 82.14 & 71.43 & 75.00 \\
vs. Deep Forcing
& 64.29 & 67.86 & 67.86 & 66.67 \\
\bottomrule
\end{tabular}
}
\end{table}

Table~\ref{tab:user-studies} reports the human preference rates of 14 participants under the 2AFC protocol. Participants compare videos generated from the same prompt and choose the better one in terms of different metrics. TetherCache is preferred over all baselines across all aspects, confirming that the improvements are perceptually noticeable beyond automatic metrics. These results suggest that TetherCache better preserves subject identity and imaging quality during extended rollouts, leading to more coherent and visually appealing long videos.

\subsection{Ablation Studies}

Table~\ref{tab:ablation-studies} validates the effectiveness of the two proposed components on the 240s generation setting. Starting from the Self-Forcing baseline, adding GRAB improves the Total Score from 71.22 to 78.80 and the Semantic Score from 44.34 to 71.05, while reducing $\Delta$ Quality Drift from 7.84 to 3.64. This indicates that relevance-diversity gated recall provides more informative long-range context than the original cache policy and substantially alleviates semantic degradation over long rollouts. Further adding TAME improves all metrics consistently, increasing the Total Score to 80.17 and further reducing $\Delta$ Quality Drift to 1.33. The additional gain confirms that the recalled memory frames are not only required to be relevant and diverse, but also need to be statistically aligned with a trusted context distribution. Overall, GRAB and TAME address the two major failure modes from complementary perspectives: GRAB mitigates information loss under a limited cache budget, whereas TAME suppresses error accumulation caused by drifted memory features. We provide a qualitative ablation study in Appendix~\ref{sec:appendix-qual-ab}.

\begin{table}[h]
\centering
\caption{\textbf{Ablation Studies.} Component ablation on 240s video generation. We progressively add GRAB and TAME to the baseline and verify their complementary effects.}
\label{tab:ablation-studies}
\resizebox{0.6\textwidth}{!}{
\begin{tabular}{l|cccc}
\toprule
\textbf{Method} & 
\thead{Total\\Score $\uparrow$} &
\thead{Quality\\Score $\uparrow$} &
\thead{Semantic\\Score $\uparrow$} &
\thead{$\Delta$ Quality\\Drift $\downarrow$} \\
\midrule
SF (Baseline) 
& 71.22 & 77.95 & 44.34 & 7.84 \\
+ GRAB 
& \underline{78.80} & \underline{80.74} & \underline{71.05} & \underline{3.64} \\
+ GRAB + TAME (\textbf{Ours}) 
& \textbf{80.17} & \textbf{81.50} & \textbf{74.85} & \textbf{1.33} \\
\bottomrule
\end{tabular}
}
\end{table}

\subsection{Analysis and Discussion}

\paragraph{Hyperparameter Sensitivity.}

Detailed results are provided in Appendix~\ref{sec:appendix-hp}. Overall, TetherCache is stable across a range of $\alpha$ and $\tau$: a moderate diversity weight ($\alpha=0.35$) best balances relevance and temporal coverage, while TAME becomes more helpful for longer rollouts and saturates around $\tau=0.6$. We use these values as defaults.

\begin{figure}[]
    \centering
    \begin{tabular}{cc}
        \includegraphics[width=0.45\linewidth]{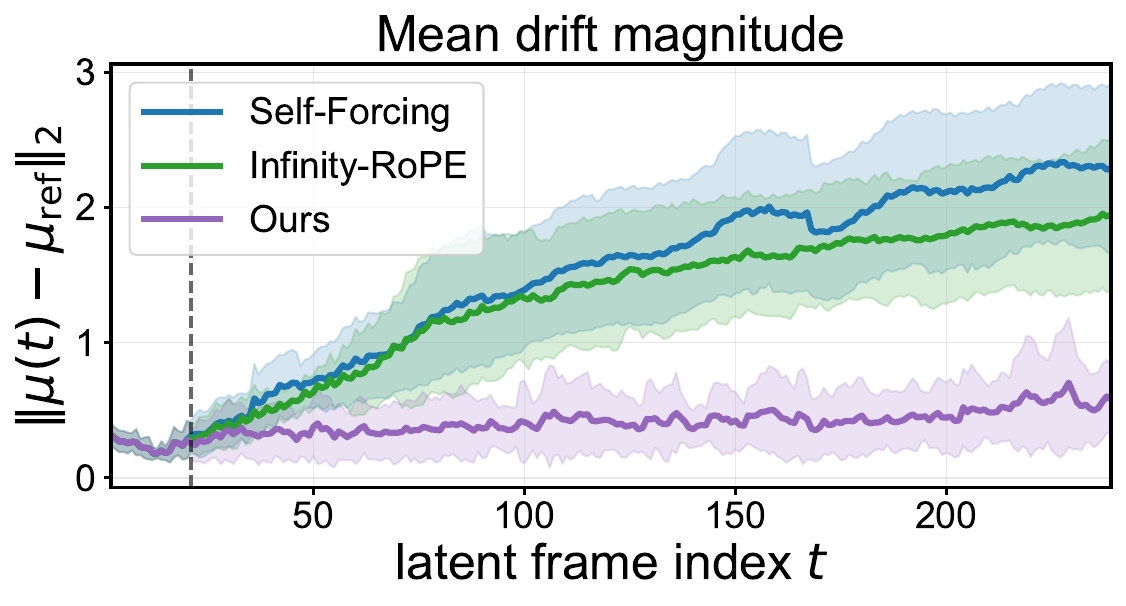} &
        \includegraphics[width=0.45\linewidth]{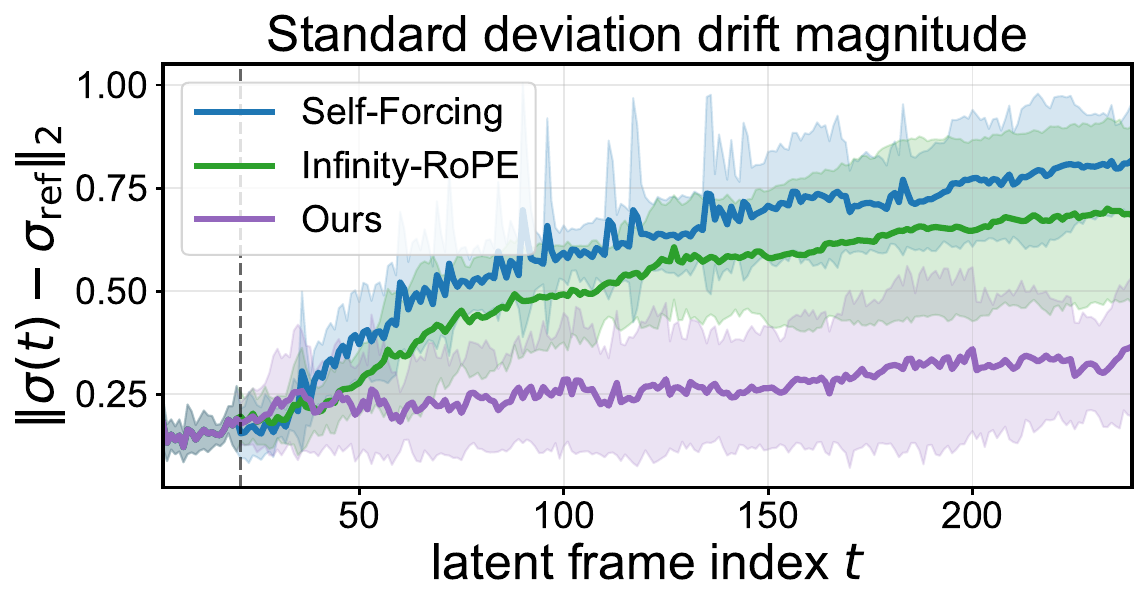} \\
        (a) & (b)
    \end{tabular}
    \caption{\textbf{Latent statistic drift analysis.} (a) Mean drift magnitude. (b) Standard deviation drift magnitude.}
    \label{fig:stat-drift-analysis}
\end{figure}

\paragraph{Latent statistic drift analysis.}
Figure~\ref{fig:stat-drift-analysis} visualizes how the latent statistics deviate from the trusted reference distribution along the AR rollout. For both the mean and standard deviation, Self-Forcing exhibits a clear monotonic increase in drift magnitude as the latent frame index grows, indicating that repeatedly conditioning on self-generated frames gradually moves the cached features away from the training-time context distribution. $\infty$-RoPE slows this process to some extent, but its latent statistics still keep drifting over long horizons. In contrast, our method keeps both mean drift and standard-deviation drift consistently low and stable after the initial context region. This directly supports the motivation of TAME: lightly aligning newly admitted memory tokens to trusted statistics prevents drifted KV features from polluting the future context, thereby reducing the accumulation of distribution mismatch during long-video generation.

\begin{figure}[h]
    \centering
    \begin{tabular}{cc}
        \includegraphics[width=0.5\linewidth]{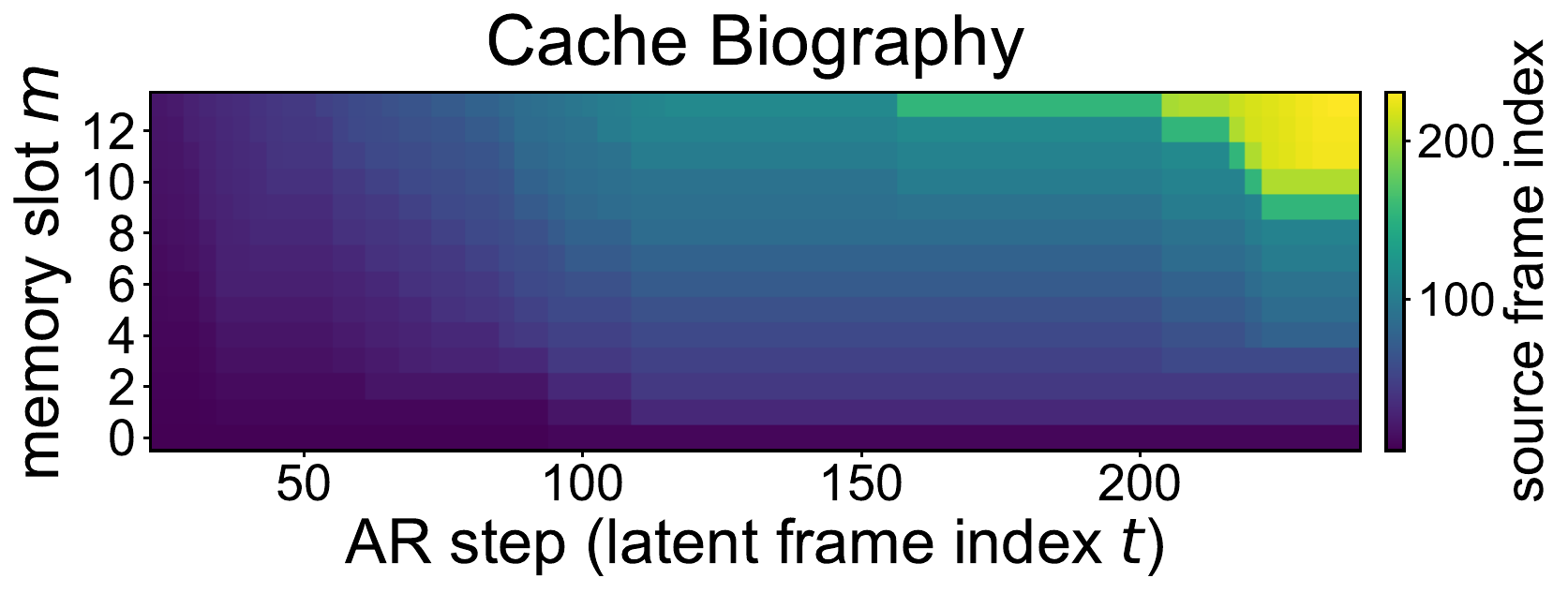} &
        \includegraphics[width=0.5\linewidth]{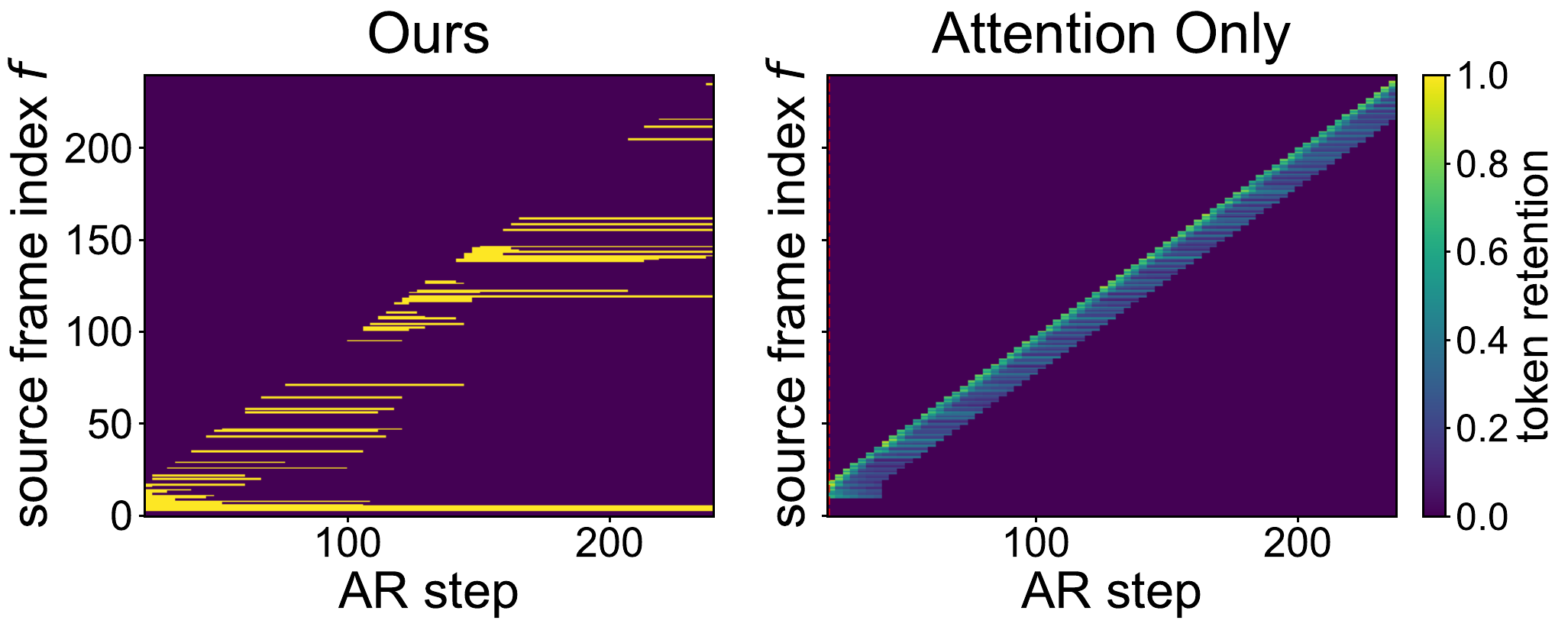} \\
        (a) & (b) \\
    \end{tabular}
    \caption{\textbf{Cache content visualization.} (a) Source-frame indices stored in each memory slot during rollout. (b) Token retention comparison between our relevance-diversity recall and attention-only selection.}
    \label{fig:cache-content-visualization}
\end{figure}

\paragraph{Cache content visualization.}
Figure~\ref{fig:cache-content-visualization} illustrates how GRAB manages the limited memory. The cache biography in Figure~\ref{fig:cache-content-visualization}(a) shows that different memory slots are assigned to source frames from different temporal ranges instead of being occupied only by the latest frames. Some slots retain early or middle-stage frames for a long period, while others are updated with more recent frames as the rollout progresses, forming a temporally broad memory layout. This behavior matches the design of GRAB, which jointly considers relevance and diversity. Figure~\ref{fig:cache-content-visualization}(b) compares our selection with an attention-only strategy. Attention-only retention concentrates around a narrow diagonal band, meaning that the cache is dominated by temporally adjacent frames and quickly forgets distant history. By contrast, our method produces sparse but long-lasting horizontal retention patterns, showing that selected historical frames can remain in memory across many AR steps. These results demonstrate that the diversity term prevents redundant local caching and enables the memory cache to provide diverse long-range context, which complements TAME's role in stabilizing the distribution of the recalled tokens.

\paragraph{Computational Overhead.}

The runtime comparison is reported in Appendix~\ref{sec:appendix-overhead}. TetherCache keeps the backbone and cache budget unchanged, adding only lightweight scoring and editing for recalled frames. It introduces $<$ 6\% latency overhead over Self-Forcing and remains comparable to baselines such as Deep Forcing, indicating a favorable quality-efficiency trade-off.
\section{Conclusion}

We present \textbf{TetherCache}, a training-free cache management strategy for stable long-horizon autoregressive video diffusion.
By combining relevance-diversity based memory selection with trusted memory editing, TetherCache preserves useful historical context while mitigating distribution drift in the KV cache.
Experiments on VBench-Long show consistent improvements over strong baselines across 30s, 60s, and 240s generation, especially in reducing long-term quality drift.
\clearpage
\bibliography{iclr2025_conference}
\bibliographystyle{iclr2025_conference}
\clearpage
\appendix
\section{Appendix}

\subsection{The TetherCache Algorithm}\label{sec:appendix-algorithm}

\begin{algorithm}[h!]
\caption{TetherCache for one attention layer}
\label{alg:tethercache}
\begin{algorithmic}[1]
\REQUIRE Cache budget $K=S+M+R$; tokens per frame $L$; GRAB weight $\alpha$; TAME strength $\tau$
\REQUIRE Current clean query $\mathbf{Q}_s$ and K/V tokens $(\mathbf{K}_s,\mathbf{V}_s)$; cache $\mathbf{S}\Vert\mathbf{M}\Vert\mathbf{R}$
\ENSURE Updated cache and attention output at rollout step $s$
\IF{cache is not filled}
    \STATE Append $(\mathbf{K}_s,\mathbf{V}_s)$ to the live cache.
    \IF{the cache first reaches $K$ frames}
        \STATE Freeze the first $S$ frames as $\mathbf{S}$.
        \STATE Initialize $\mathbf{M}$ with the next $M$ frames and store their global indices $g_c$.
        \STATE Set the last $R$ frames as the recent window $\mathbf{R}$.
    \ENDIF
    \STATE Apply block-relative RoPE on cached keys and return attention.
\ENDIF
\STATE Snapshot the front recent frame $e_s\leftarrow\operatorname{head}(\mathbf{R})$.
\STATE Roll $\mathbf{R}$ left and write $(\mathbf{K}_s,\mathbf{V}_s)$ to its tail.
\STATE Form $\mathcal{P}_s\leftarrow\mathbf{M}_{s}^{\mathrm{cur}}\cup\{e_s\}$.
\STATE $\sigma_s\leftarrow\max(1,\frac{1}{2}(\max_{c\in\mathcal{P}_s}g_c-\min_{c\in\mathcal{P}_s}g_c+1))$.
\FORALL{$c\in\mathcal{P}_s$}
    \STATE $\ell(c)\leftarrow\frac{1}{HL^2}\sum_{h,q,k}\langle\mathbf{Q}_{s,q}^{h},\mathbf{K}_{c,k}^{h}\rangle/\sqrt{D}$.
\ENDFOR
\STATE $\phi^{\mathrm{imp}}(c)\leftarrow\exp(\ell(c))/\sum_{c'\in\mathcal{P}_s}\exp(\ell(c'))$ for all $c$.
\FORALL{$c\in\mathcal{P}_s$}
    \STATE $r(c)\leftarrow\max_{c'\ne c}\exp(-|g_c-g_{c'}|/\sigma_s)\phi^{\mathrm{imp}}(c')$.
    \STATE $\phi(c)\leftarrow\phi^{\mathrm{imp}}(c)+\alpha\max(0,1-r(c))$.
\ENDFOR
\STATE $\mathbf{M}_{s+1}\leftarrow\operatorname{TopM}_{c\in\mathcal{P}_s}\phi(c)$.
\IF{$e_s\in\mathbf{M}_{s+1}$}
    \STATE $\mathcal{T}_s\leftarrow\mathbf{S}\cup\mathbf{M}_{s}^{\mathrm{cur}}$.
    \FORALL{$\mathbf{x}\in\{\mathbf{K}_{e_s},\mathbf{V}_{e_s}\}$}
        \STATE $(\mu_{\mathbf{x}},\sigma_{\mathbf{x}})\leftarrow\operatorname{Stat}(\mathbf{x})$.
        \STATE $(\mu_{\mathcal{T}},\sigma_{\mathcal{T}})\leftarrow\operatorname{Stat}(\mathcal{T}_s)$ using the corresponding K/V trusted tokens.
        \STATE $\tilde{\mathbf{x}}\leftarrow\sigma_{\mathcal{T}}\odot(\mathbf{x}-\mu_{\mathbf{x}})/\sigma_{\mathbf{x}}+\mu_{\mathcal{T}}$.
        \STATE $\hat{\mathbf{x}}\leftarrow(1-\tau)\mathbf{x}+\tau\tilde{\mathbf{x}}$.
    \ENDFOR
    \STATE Replace $e_s$ in $\mathbf{M}_{s+1}$ with $(\hat{\mathbf{K}}_{e_s},\hat{\mathbf{V}}_{e_s})$.
\ENDIF
\STATE Rewrite the physical memory region with $\mathbf{M}_{s+1}$ and update global-index metadata.
\STATE Apply block-relative RoPE on cached keys and return attention over $\mathbf{S}\Vert\mathbf{M}_{s+1}\Vert\mathbf{R}$.
\end{algorithmic}
\end{algorithm}

Algorithm~\ref{alg:tethercache} summarizes the layer-wise procedure of TetherCache. The implementation follows the same fixed cache layout in Eq.~\ref{eq:tether-layout}: newly generated K/V tokens are written into the recent region, the evicted recent frame competes with existing memory through GRAB, and TAME is applied only when the evicted frame is newly admitted. In AR diffusion inference, the same slot is overwritten during noisy denoising passes, while the GRAB/TAME update is executed on the final clean context pass using the clean query tokens $\mathbf{Q}_s$.

\subsection{Qualitative Ablation Study}\label{sec:appendix-qual-ab}

\begin{figure}[h]
    \centering
        \includegraphics[width=0.9\linewidth]{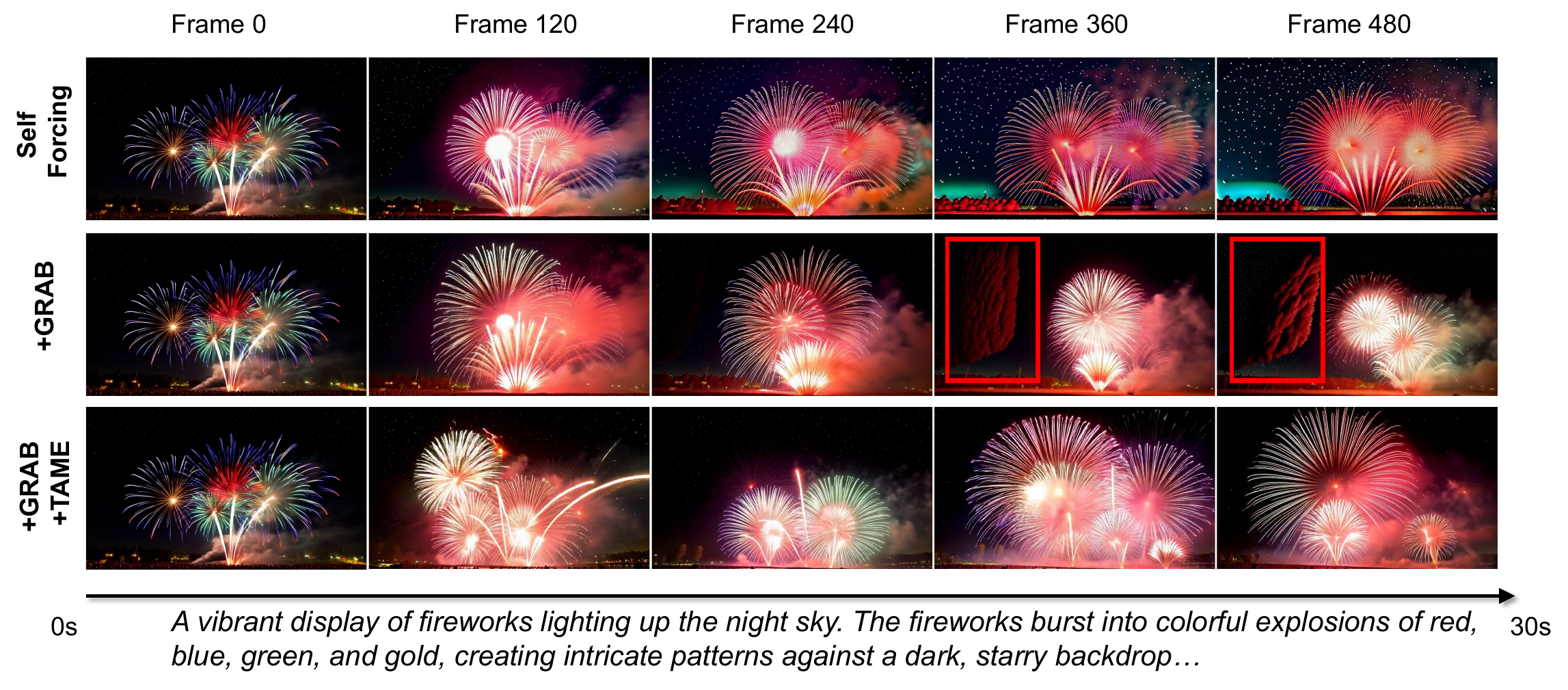}
    \caption{\textbf{Qualitative ablation results.} }
    \label{fig:ablation-qual}
\end{figure}

Figure~\ref{fig:ablation-qual} presents the qualitative results of the ablation study. In 30-second video generation, the baseline method quickly exhibits significant color shifts and inconsistencies. After applying GRAB, the color consistency of the generated video is substantially improved, although artifacts highlighted by the red boxes still appear. When both GRAB and TAME are applied simultaneously, these artifacts are eliminated, enabling stable and consistent long-video generation.

\subsection{Hyperparameter Sensitivity}\label{sec:appendix-hp}

\begin{figure}[h!]
    \centering
    \begin{tabular}{cc}
        \includegraphics[width=0.45\linewidth]{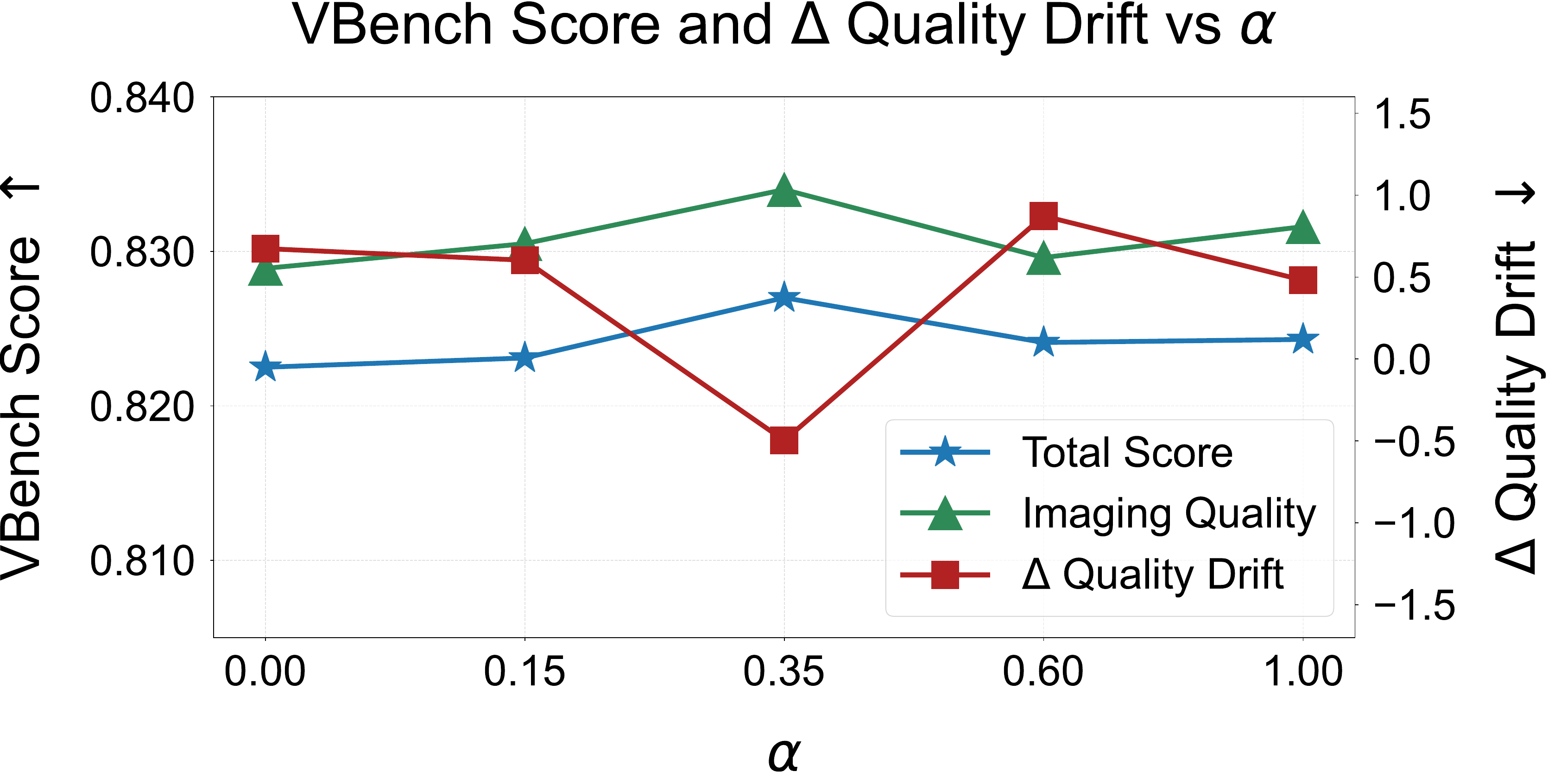} &
        \includegraphics[width=0.45\linewidth]{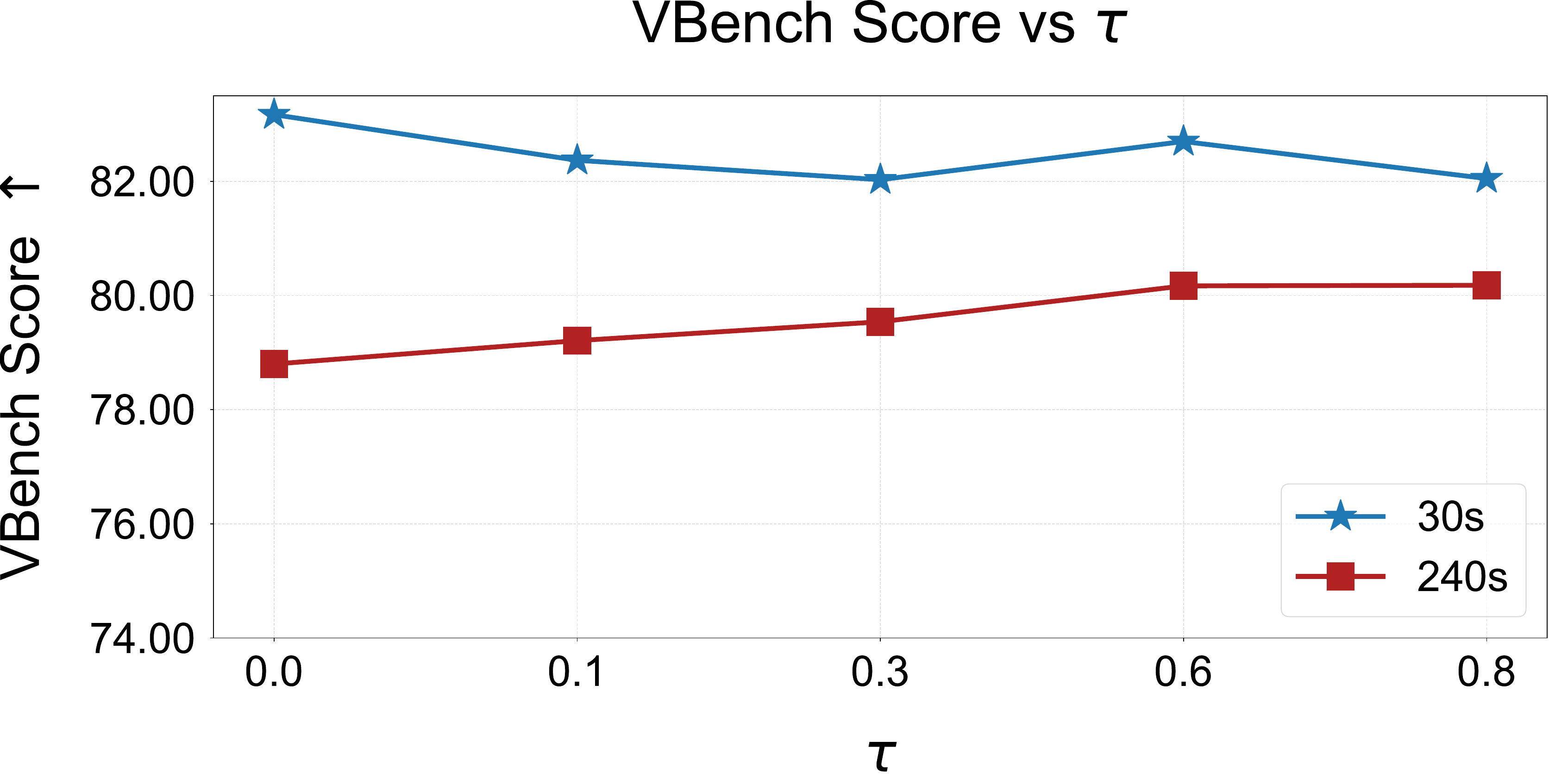} \\
        (a) & (b)
    \end{tabular}
    \caption{\textbf{Hyperparameter sensitivity analysis.}
    (a) VBench score and $\Delta$ Quality Drift under different $\alpha$s.
    (b) VBench score under different $\tau$s.}
    \label{fig:hyperparameter-analysis}
\end{figure}

Figure~\ref{fig:hyperparameter-analysis} studies the sensitivity of TetherCache to the two key hyperparameters: the diversity weight $\alpha$ in GRAB and the editing strength $\tau$ in TAME. As shown in Figure~\ref{fig:hyperparameter-analysis}(a), the overall performance is relatively stable across different $\alpha$ values, indicating that GRAB is not overly sensitive to the exact balance between attention-based relevance and temporal diversity. Nevertheless, setting $\alpha=0.35$ achieves the best Total Score and Imaging Quality, while also producing the lowest $\Delta$ Quality Drift. When $\alpha$ is too small, the selection is dominated by attention relevance and tends to keep temporally clustered frames, which weakens long-range coverage. When $\alpha$ becomes too large, the cache may over-emphasize temporal dispersion and admit less relevant frames. The intermediate value therefore provides a better trade-off between recalling useful context and maintaining diverse historical coverage.

Figure~\ref{fig:hyperparameter-analysis}(b) analyzes the effect of the TAME strength $\tau$. For short 30s generation, the VBench score changes only mildly with $\tau$, suggesting that limited rollout length introduces relatively small distribution drift and thus requires less aggressive correction. In contrast, for 240s generation, increasing $\tau$ consistently improves the score until it saturates around $\tau=0.6$, demonstrating that statistical alignment becomes increasingly important as the autoregressive rollout becomes longer. A very large $\tau$ brings no further gain, as excessive editing may suppress useful content-specific variations in the admitted memory tokens. We therefore use $\alpha=0.35$ and $\tau=0.6$ as the default configuration, which offers a strong and robust balance across both short and long generation settings.

\begin{figure}[h!]
    \centering
    \begin{tabular}{cc}
        \includegraphics[width=0.45\linewidth]{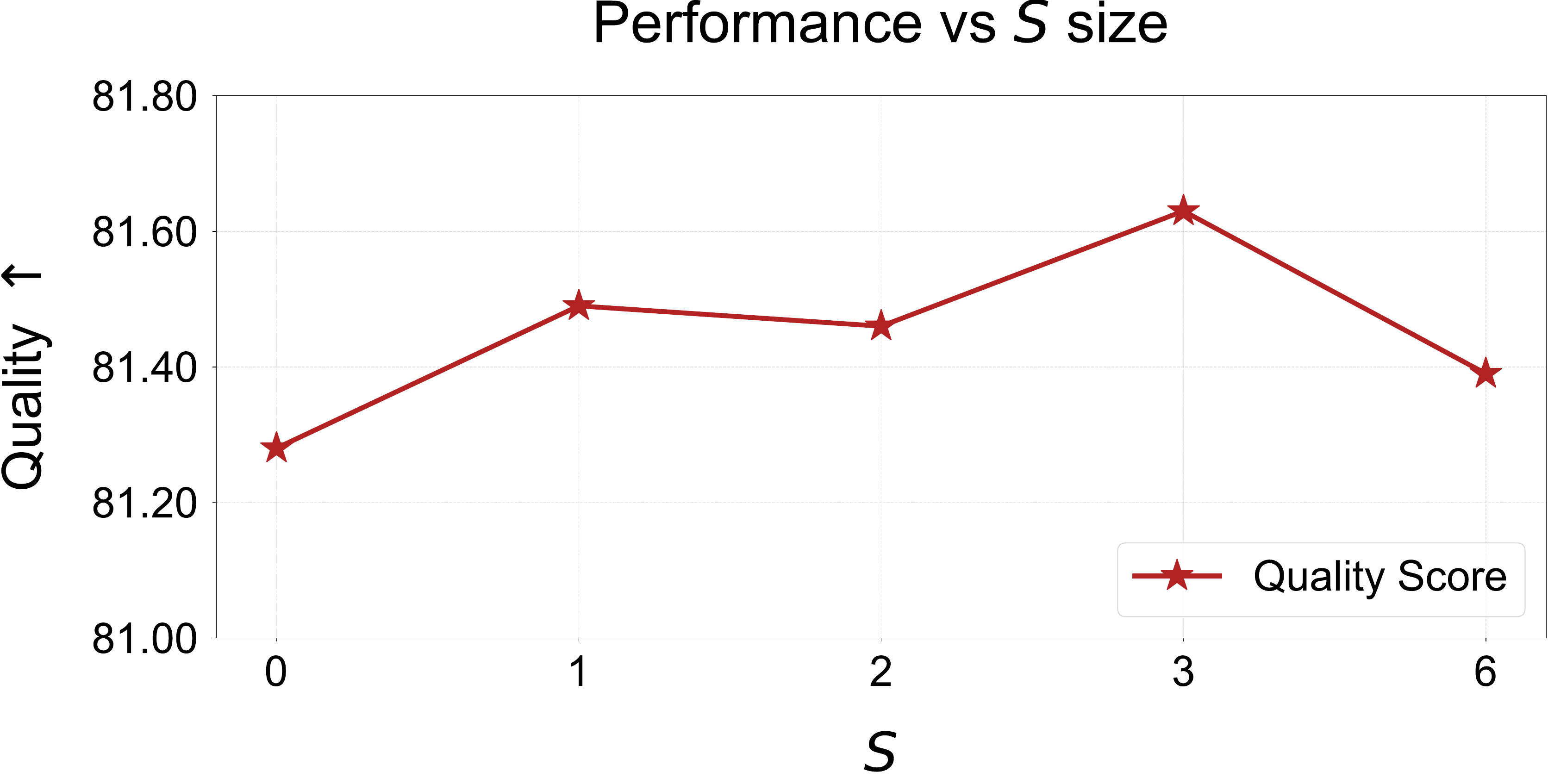} &
        \includegraphics[width=0.45\linewidth]{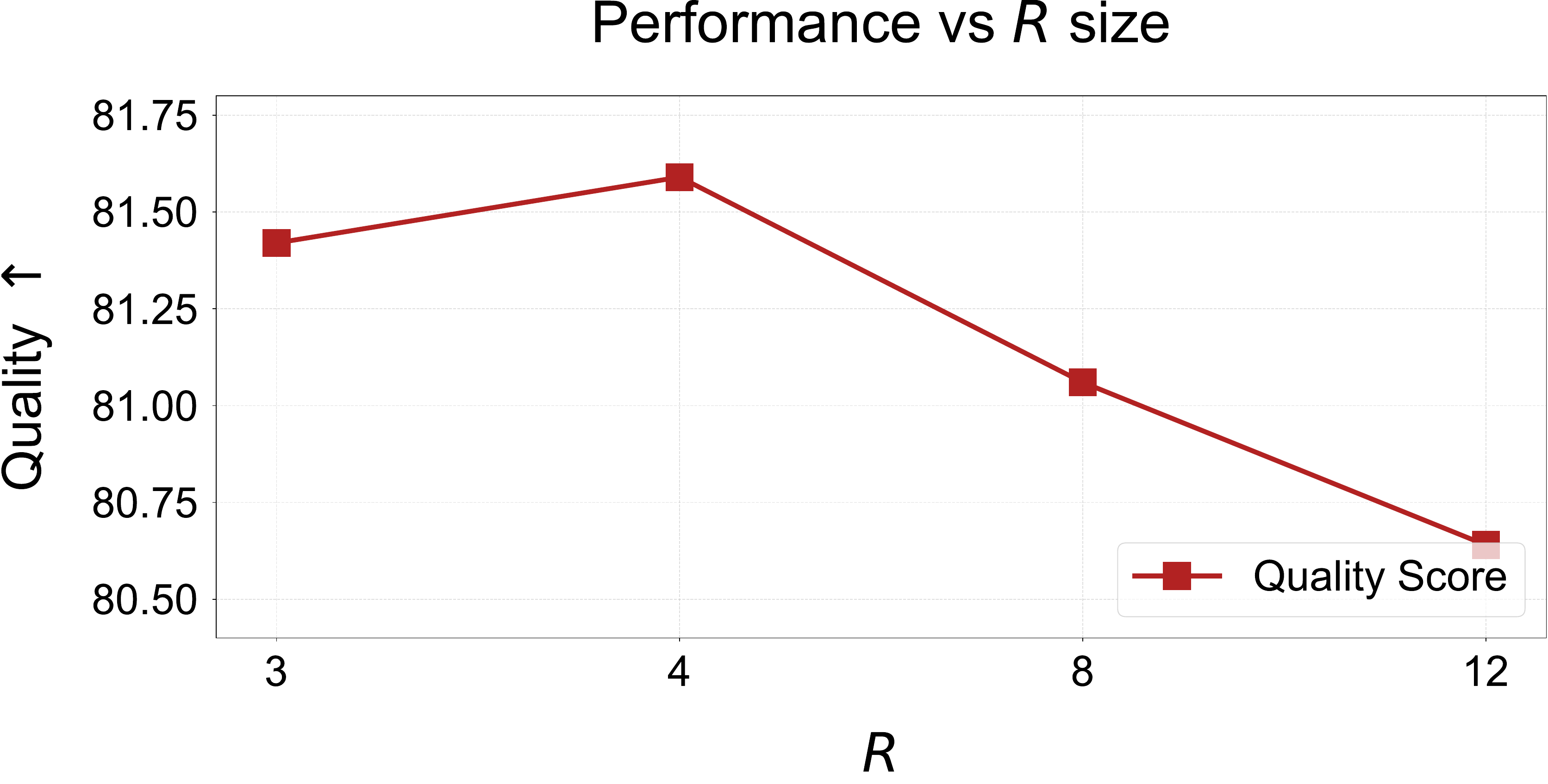} \\
        (a) & (b)
    \end{tabular}
    \caption{\textbf{Cache budget analysis.}
    (a) The impact of sink size $S$ on model performance.
    (b) The impact of recent size $R$ on model performance.}
    \label{fig:cache-size-analysis}
\end{figure}

Figure~\ref{fig:cache-size-analysis} studies cache allocation under a fixed total budget. Performance improves when increasing the sink size from $S=0$ to $S=3$, confirming the value of trusted sink statistics, but drops at $S=6$ as too many sink slots reduce memory capacity. For the recent window, $R=4$ performs best. Larger $R$ weakens long-range memory by occupying more cache slots. These results support our default setting $S=3$ and $R=4$, which balances trusted priors, local continuity, and selective long-range recall.

\subsection{Computational Overhead}\label{sec:appendix-overhead}

We analyze the computational overhead of TetherCache. The FLOPs of the baseline method’s regular computation for each frame and each self-attention layer are estimated as follows (chunk-wise generation, 3 frames per chunk):
\begin{equation}
\underbrace{\left(T+1\right)}_{\rm{timesteps}}\left(\underbrace{18L\left(HD\right)^2}_{\mathrm{QKV}\ \rm{projection}} + \underbrace{12HDKL^2}_{\rm{Softmax}\left(\mathbf{QK^\top}\right)\mathbf{V}} + \underbrace{6LD^2}_{\rm{Output}\ \rm{projection}}\right).
\end{equation}

The additional computational cost introduced by GRAB and TAME is:
\begin{equation}
    \underbrace{6HDL^2\left(M+3\right)}_{\rm{GRAB}} + \underbrace{\left(S+M+2\right)HDL}_{\rm{TAME}}.
\end{equation}

From the two equations above, we can see that denoising each chunk and saving the cache require $T+1$ forward passes, whereas GRAB computes the Attention Mass Score only once among them, which greatly reduces the proportion of computational overhead introduced by GRAB. Although the overhead scales quadratically with $L$, $L$ only denotes the token number in a single frame, which is constant and therefore does not become a scalability bottleneck.

\begin{table}[h]
\centering
\caption{\textbf{Computational overhead results.} All metrics are measured on one NVIDIA H20 (96GB) with Flash Attention~\citep{dao2023flashattention2fasterattentionbetter}.}
\label{tab:computational-overhead}
\resizebox{0.75\textwidth}{!}{
\begin{tabular}{l|ccc}
\toprule
\textbf{Method} & 
\thead{\# Params} &
\thead{Latency (s) $\downarrow$} &
\thead{Throughput (FPS) $\uparrow$} \\
\midrule
Self Forcing (Baseline)
& 1.3B & 1126.1 & 3.41 \\
MemRoPE
& 1.3B & 1211.4 & 3.17  \\
Deep Forcing
& 1.3B & 1185.2 & 3.24  \\
Ours
& 1.3B & 1192.5 & 3.22  \\
\bottomrule
\end{tabular}
}
\end{table}

Table~\ref{tab:computational-overhead} compares the inference efficiency of TetherCache with representative baselines for 240-second video generation. Since TetherCache is training-free and does not modify the backbone network, it keeps the same number of parameters as Self-Forcing. Its additional cost mainly comes from the lightweight GRAB candidate scoring and the TAME statistics alignment applied only when a frame is admitted into memory. Compared with the Self-Forcing baseline, TetherCache increases latency from 1126.1s to 1192.5s, corresponding to less than 6\% overhead, while maintaining a practical throughput of 3.22 FPS. This overhead is modest given the substantial gains in generation quality and drift reduction reported in Table~\ref{tab:main-results}.

Compared with other long-context baselines, TetherCache has comparable runtime efficiency: it is faster than MemRoPE and close to Deep Forcing in both latency and throughput. These results indicate that the proposed cache management operations do not introduce a heavy computational burden. Instead, TetherCache achieves a favorable quality-efficiency trade-off by reusing the original KV-cache structure and adding only small per-frame metadata updates, candidate scoring, and token-statistic normalization.

\end{document}